**Automated Body Composition Analysis Using DAFS Express on 2D MRI Slices at L3 Vertebral Level**

Varun Akella[1,2], Razeyeh Bagherinasab[1], Jia Ming Li[1], Long Nguyen[1], Vincent Tze Yang Chow[1], Hyunwoo Lee[4], Karteek Popuri[3], Mirza Faisal Beg[1]

[1] School of Engineering Science, Simon Fraser University, Vancouver, Canada

[2] Department of Biomedical Physiology and Kinesiology, Simon Fraser University, Vancouver, Canada

[3] Department of Computer Science, Memorial University of Newfoundland, St. John's, Canada

[4] Division of Neurology, Department of Medicine, University of British Columbia, Vancouver, Canada

**Keywords:** body composition, automated 2D segmentation, artificial intelligence, magnetic resonance imaging

**ABSTRACT**

**Background**
Body composition analysis is vital in assessing health conditions such as obesity, sarcopenia, and metabolic syndromes. MRI provides detailed images of skeletal muscle (SKM), visceral adipose tissue (VAT), and subcutaneous adipose tissue (SAT), but their manual segmentation is labor-intensive and limits clinical applicability. This study validates an automated tool for MRI-based 2D body composition analysis- (Data Analysis Facilitation Suite (DAFS) Express), comparing its automated measurements with expert manual segmentations using UK Biobank data.

**Methods**
A cohort of 399 participants from the UK Biobank dataset was selected, yielding 423 single L3 slices for analysis. DAFS Express performed automated segmentations of SKM, VAT, and SAT, which were then manually corrected by expert raters for validation. Evaluation metrics included Jaccard coefficients, Dice scores, Intraclass Correlation Coefficients (ICCs), and Bland-Altman Plots to assess segmentation agreement and reliability.

**Results**
High agreements were observed between automated and manual segmentations with mean Jaccard scores: SKM 99.03%, VAT 95.25%, and SAT 99.57%; and mean Dice scores: SKM 99.51%, VAT 97.41%, and SAT 99.78%. Cross-sectional area comparisons showed consistent measurements with automated methods closely matching manual measurements for SKM and SAT, and slightly higher values for VAT (SKM: Auto 132.51 cm², Manual 132.36 cm²; VAT: Auto 137.07 cm², Manual 134.46 cm²; SAT: Auto 203.39 cm², Manual 202.85 cm²). ICCs confirmed strong reliability (SKM: 0.998, VAT: 0.994, SAT: 0.994). Bland-Altman plots revealed minimal biases, and boxplots illustrated distribution similarities across SKM, VAT, and SAT areas. On average DAFS Express took 18 seconds per DICOM for a total of 126.9 minutes for 423 images to output segmentations and measurement PDF's per DICOM.

**Conclusion**
Automated segmentation of SKM, VAT, and SAT from 2D MRI images using DAFS Express showed comparable accuracy to manual segmentation. This underscores its potential to streamline image analysis processes in research and clinical settings, enhancing diagnostic accuracy and efficiency. Future work should focus on further validation across diverse clinical applications and imaging conditions.

**Introduction**

Body composition analysis is a crucial component in understanding various health conditions, including obesity, sarcopenia, and metabolic syndromes. Accurate quantification of skeletal muscle (SKM), visceral adipose tissue (VAT), and subcutaneous adipose tissue (SAT) from medical imaging such as magnetic resonance imaging (MRI) provides valuable insights into patient health and disease progression [1–3]. However, manual segmentation of these tissues from MRI scans is a labor-intensive and time-consuming process, which can limit its widespread clinical application.

Advancements in artificial intelligence (AI) and machine learning have led to the development of automated tools for medical image analysis [4]. One such tool is the Data Analysis Facilitation Suite (DAFS) Express, a software solution designed to automate 2D body composition measurements from CT and MRI scans.

DAFS Express is particularly focused on analyzing body composition from In-Phase Dixon MRI sequences, a commonly used imaging technique for fat and muscle tissues. By automating the segmentation process, DAFS Express has the potential to significantly reduce the workload on radiologists and clinicians, allowing for more efficient and accurate assessments [4].

The primary objective of this study was to validate the accuracy of DAFS Express in automating body composition measurements at the L3 vertebral level from MRI scans. The L3 level is often selected for body composition analysis due to its strong correlation with whole-body adipose and muscle tissue distribution [5,6]. We aim to compare the automated measurements generated by DAFS Express against manual segmentations performed by expert raters. The accuracy of the segmentation outputs were evaluated using Jaccard coefficients and Dice scores, which are standard metrics for assessing the overlap between two sets of segmentations.

**Methods**

*Participant Population*

A cohort of 399 participants was selected from the open source UK Biobank dataset, which is a comprehensive biomedical database of over 500,000 participants from the United Kingdom (UK) [7–10]. Each selected participant had undergone MRI imaging with their L3 vertebral level visible. From these participants, a total of 423 single L3 slices were analyzed, as some participants had multiple scans at different timestamps. This allowed for the assessment of both inter- and intra- participant variability in the segmentation process. MRI scans that did not include the L3 vertebral level were excluded from the study. *Table 1* displays the population metrics.

The selection process involved randomly choosing Electronic Identification Numbers (EIDs) from a CSV file containing all UK Biobank participant data [10,11]. Participants were included if their MRI scans had the L3 vertebral visible within the scan. Exclusion criteria included scans without a discernible L3 level.

| Table 1: Cohort Characteristics | |
|---|---|
| Gender | |
|   Male | 179 (44.9%) |
|   Female | 220 (55.1%) |
| Birth Year (median, years) | 1951 [1936 – 1969] |
| Manual analysis (mean, cm$^2$) | |
|   SKM | 132.36 |
|   SAT | 202.85 |
|   VAT | 134.46 |
| Automated analysis (mean, cm$^2$) | |
|   SKM | 132.51 |
|   SAT | 203.39 |
|   VAT | 137.07 |

**Table 1** Table showing population characteristics such as gender distribution, median birth years, and mean cross sectional areas for SKM, SAT, and VAT for both manual analysis and automatic analysis.

*Automated Segmentation*

The MRI scans were processed using the latest version of DAFS Express. The automated segmentation process involved several key steps: *Extraction* - For each of the 423 series of DICOM images, a single L3 slice was manually extracted by a rater proficient in anatomical analysis using DAFS Express. The rater utilized their expertise to identify the L3 vertebrae slice accurately. *Segmentation* - DAFS Express then automatically segmented the extracted L3 level images. The software identified and measured the cross-sectional areas of SKM, SAT, and VAT. The segmentation algorithm processed the images without any manual intervention, ensuring the consistency and objectivity of the automated measurements. *Quantification* - The software quantified the segmented areas, providing measurements of cross-sectional areas (cm²) and mean intensity for SKM, SAT, and VAT.

*Manual Segmentation*

Following the automated segmentation by DAFS Express, expert raters reviewed and manually corrected any errors using the same software. The manual correction process involved the following steps: *Review* - The automatically segmented L3 slices were randomly assigned to raters with significant experience in anatomical segmentation. Each rater reviewed the automated segmentations of SKM, SAT, and VAT. *Correction* - Using the tools provided in DAFS Express, the raters manually adjusted the segmentation boundaries to correct any inaccuracies. This step involved precise adjustments to ensure the boundaries matched the anatomical structures accurately. The raters made corrections based on visual assessment and their expertise. *Recording* - After corrections were made, the raters recorded the cross-sectional areas and mean intensity values for each tissue type from the software-generated reports. These manually corrected segmentations were considered the reference standard for the study. The corrected segmentations were saved and stored for further analysis.

*Accuracy Assessment*

The accuracy of DAFS Express was assessed by comparing the automated segmentations to the manually corrected segmentations. The assessment process included the following steps: *Conversion* - Individual segmentation files were converted to tag files using a proprietary script developed in-house. This script ensured that the data was in a format suitable for detailed comparison and analysis. *Metrics Calculation* - A separate proprietary script was utilized to calculate Jaccard coefficients and Dice scores, which are standard metrics for evaluating the overlap between two sets of segmentations [12,13]. These metrics provide a quantitative measure of the agreement between the manual and automated segmentations. *Jaccard Coefficient* is defined as the intersection over union of the manual and automated segmentations. It quantifies the common area divided by the total area covered by both segmentations. *Dice Scores* are defined as twice the intersection divided by the sum of the manual and automated segmentations. It provides a measure of similarity, with higher values indicating better agreement. The results from these metrics provided a comprehensive assessment of the performance of DAFS Express in automating body composition measurements from MRI scans.

*Statistical Analysis*

The following statistical analyses were conducted to evaluate the agreement between automated and manual segmentations of subcutaneous skeletal muscle (SKM), visceral adipose tissue (VAT), and subcutaneous adipose tissue (SAT). Intraclass Correlation Coefficients (ICCs) were calculated to evaluate the reliability and agreement between automated and manual segmentations for each tissue type. ICC values close to 1 indicate high agreement [14]. Bland-Altman Plots were utilized to visually inspect the agreement between automated and manual measurements [15]. These plots depict the mean difference between methods against their average, highlighting any systematic biases.

*Evaluation in Challenging Imaging Conditions*

To ensure the accuracy of segmentation under challenging conditions, our study included images of varying quality, some of which were suboptimal due to inherent issues in the imaging process. These issues included images in the middle of being stitched together, participant movement during imaging, which resulted in blurred areas as well as poor image quality. In such cases, the rater spent additional time reviewing the segmented images to verify that the software correctly identified and segmented the relevant regions. This extra scrutiny was crucial for maintaining the integrity of our analysis and ensuring that the software's performance was accurately assessed.

**Results**

*Segmentation Accuracy Assessment*

The Jaccard and Dice scores provide measures of overlap and agreement between automated and manual segmentations, indicating high similarity across skeletal muscle (SKM), visceral adipose tissue (VAT), and subcutaneous adipose tissue (SAT) areas (SKM: Jaccard 99.03, Dice 99.51; VAT: Jaccard 95.25, Dice 97.41; SAT: Jaccard 99.57, Dice 99.78). These metrics demonstrate robust agreement between methods in delineating tissue boundaries, supported by visual comparisons of raw segmentations. DAFS Express was able to achieve high scores comparable to manual segmentation but at a lower time cost. On average DAFS Express took 18 seconds per DICOM for a total of 126.9 minutes for 423 images to output segmentations and measurement PDF's per DICOM.

*Cross-sectional Area Comparison*

Comparison of cross-sectional areas showed consistent measurements between automated and manual methods. Mean areas were comparable for SKM (Auto: 132.51 cm², Manual: 132.36cm²) and SAT (Auto: 203.39 cm², Manual: 202.85 cm²), with slightly higher automated area for VAT (Auto: 137.07 cm², Manual: 134.46 cm²). This highlights the effectiveness of automated segmentation in capturing tissue boundaries akin to manual measurements. *Fig. 1a-h* illustrates some examples of the manual and automatic segmentations from 8 randomly chosen participants from the cohort.

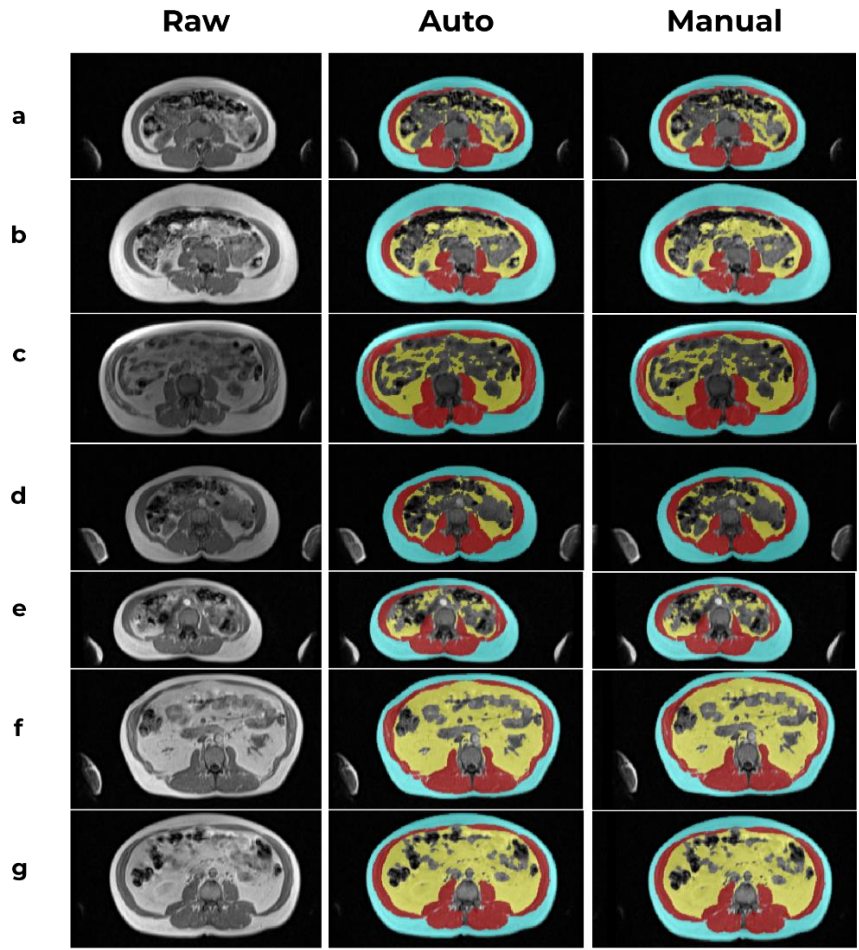

**Fig. 1** Comparison of original DICOM images with manual and automated segmentation of body composition from MRI scans. Blue represents subcutaneous adipose tissue (SAT), red represents skeletal muscle mass (SKM), and yellow represents visceral adipose tissue (VAT). Image (a), (b), (c), (d), (e), (f), and (g) all represent different participants at the L3 vertebral level.

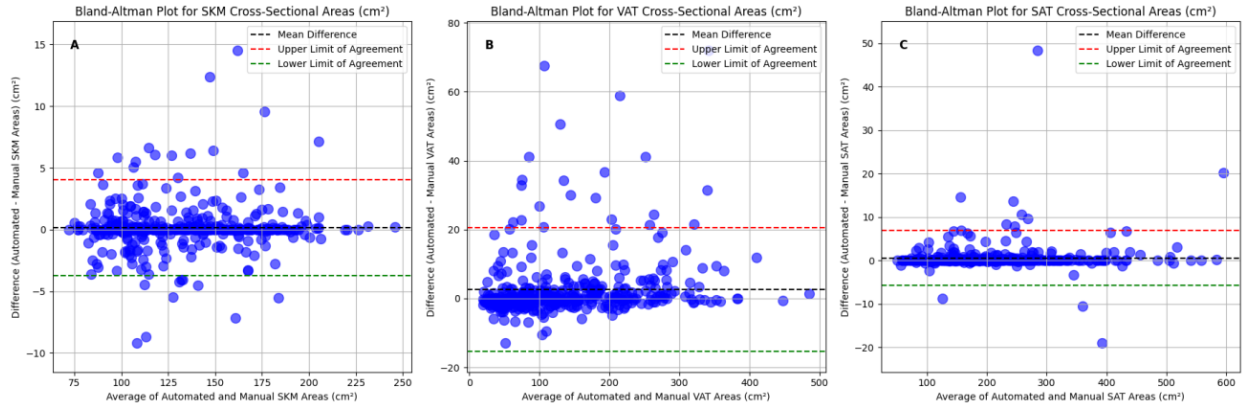

**Fig. 2** Combined Bland-Altman Plots for Cross-Sectional Areas of SKM (A), VAT (B), and SAT (C). The plots show the agreement between automated and manual measurements of SKM, VAT, and SAT areas. The dashed lines represent the mean difference between the automated and manual measurements, and the dotted lines represent the upper and lower limits of agreement (mean difference ± 1.96 standard deviations). The scatter plots show individual data points, with the x-axis representing the average of the automated and manual measurements, and the y-axis representing the difference between the automated and manual measurements. The horizontal lines in the plots indicate the limits of agreement.

Bland-Altman plots (*Fig.2*) visually depict agreement between automated and manual segmentations, showing the difference versus the average of the two methods. Calculated limits of agreement were (lower bound – upper bound) - *SKM area: 3.73 - 4.03, VAT area: -15.30 - 20.51, and SAT area: -5.83 - 6.91.* The Bland-Altman plots demonstrated narrow limits of agreement due to significant data points falling within the limits of agreement for all tissues, indicating strong agreement and low variability between the automated DAFS Express and manual segmentations. The consistent scatter of data points around the mean difference line suggests reliable performance across the range of measurements, with few outliers primarily due to poor image quality or anatomical variations.

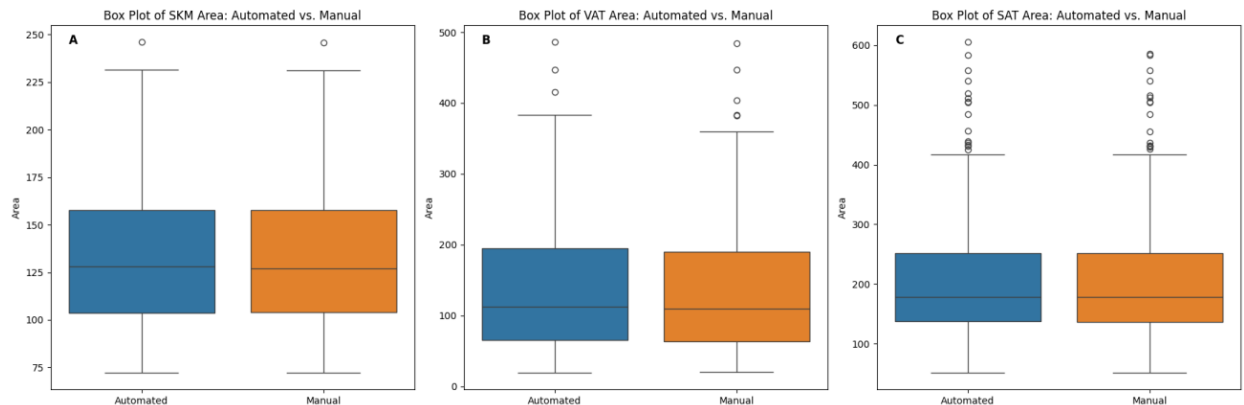

**Fig. 3** Combined Boxplots of Cross-Sectional Areas of SKM, VAT, and SAT. The boxplots compare the distribution of automated and manual measurements of SKM, VAT, and SAT areas. The boxplots show the median (middle line), the interquartile range (box), and the minimum and maximum values (whiskers). The individual data points are plotted as dots. The subplots are labeled A, B, and C to correspond to the SKM, VAT, and SAT areas, respectively.

Box plots (*Fig. 3*) illustrate the distribution of cross-sectional areas which offers comprehensive insights into segmentation accuracy and variability across tissues. These boxplots also show significant overlap between the means of automated and manual cross-sectional areas for SKM (*Fig. 3A*), VAT (*Fig. 3B*), and SAT (*Fig. 3C*). The spread of the data points between automated and manual segmentations are similar for each tissue.

*Reliability Analysis*

High Intraclass Correlation Coefficient (ICC) values (SKM: 0.998, VAT: 0.994, SAT: 0.994) underscore the reliability and reproducibility of automated segmentation compared to manual methods. These coefficients reflect strong agreement [15] across different tissue types, affirming the consistency in segmentation outcomes. *Fig. 4* showcases more participants, visually there is little to no difference between manual and automatic segmentations as raters agreed with the automatic segmentation most of the time.

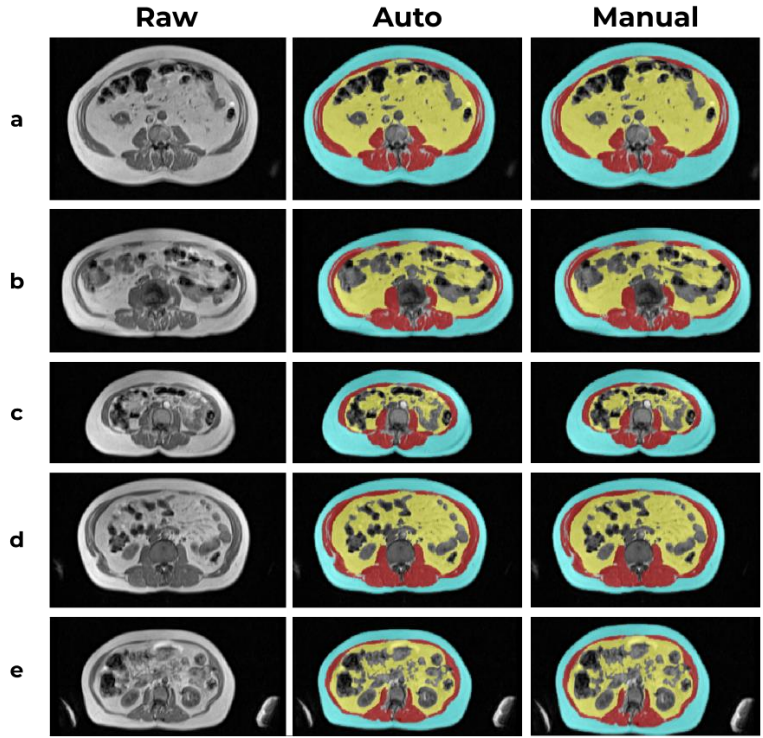

**Fig. 4** Comparison of original DICOM images with manual and automated segmentation of body composition from MRI scans. Blue represents subcutaneous adipose tissue (SAT), red represents skeletal muscle mass (SKM), and yellow represents visceral adipose tissue (VAT). Images (a), (b), (c), (d), and (e) all represent different participants.

*Performance in Challenging Imaging Conditions*

Despite the presence of challenging conditions, such as blurring from image stitching processes, motion artifacts, and low resolution, DAFS Express still performed well. One example is shown in Figure 5, where the image quality was compromised. The Jaccard scores for SAT and VAT were, 97.28 and 85.87 (*Fig. 5a*), 97.81 and 93.29 (*Fig. 5b*), 96.194 and 100 (*Fig. 5c*). Such scores were maintained consistently throughout lower quality images. The software accurately segmented the relevant anatomical regions, maintaining high accuracy despite the suboptimal image quality. This result highlights the software's capability to deliver reliable segmentations even when image quality is less than ideal, underscoring its potential applicability in clinical settings where imaging conditions can vary significantly.

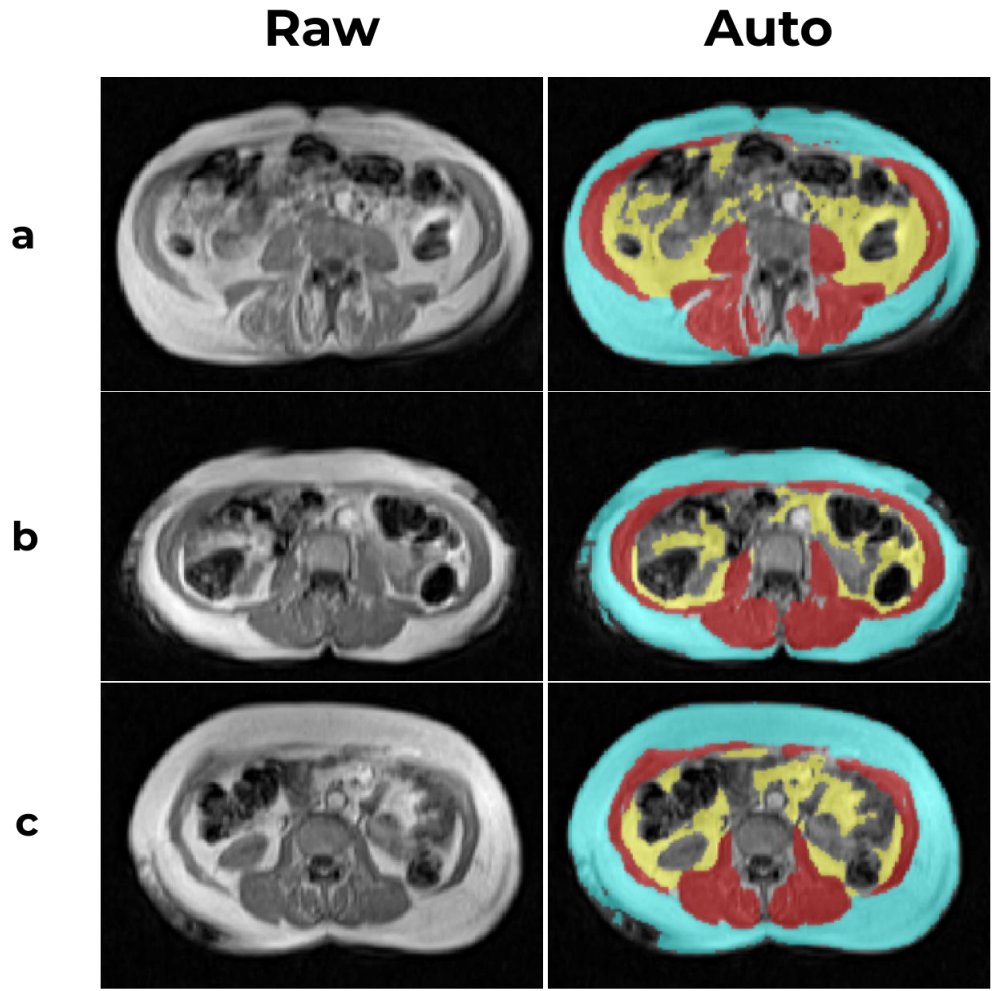

**Fig. 5** Examples of DAFS Express automatic segmentation on lower quality MRI images. The left panel shows the original MRI image with suboptimal quality, and the right panel shows the corresponding segmentation output by DAFS Express.

**Discussion**

In this study, we conducted a comprehensive evaluation of automated versus manual segmentations of subcutaneous skeletal muscle (SKM), visceral adipose tissue (VAT), and subcutaneous adipose tissue (SAT) areas from 2D MRI slices. Our findings provide valuable insights into the accuracy and reliability of automated segmentation tools compared to manual methods.

*Accuracy and Agreement Metrics*

Our analysis of mean Jaccard and Dice scores across SKM, VAT, and SAT areas demonstrates robust agreement between automated and manual segmentations. The mean Jaccard scores ranged from 99.03% to 99.57%, while the mean Dice scores ranged from 99.51% to 99.78%. These metrics indicate substantial overlap and similarity between segmentation methods, affirming the software's capability to accurately delineate tissue boundaries.

*Statistical Analysis and Reliability*

The Bland-Altman plots provided further insights into the agreement between automated and manual segmentations. These plots revealed narrow limits of agreement, suggesting consistent performance across varying tissue areas. The Intraclass Correlation Coefficients (ICCs) reinforced these findings with values exceeding 0.99 for VAT and SKM, and SAT. High ICC values indicate strong agreement and reliability between the automated software and manual segmentation, essential for clinical and research applications [14].

*Performance in Challenging Imaging Conditions*

A notable strength of the DAFS Express software was its performance in lower quality image conditions. Our study included MRI images that exhibited suboptimal quality due to factors such as blurring from image stitching processes, motion artifacts, and low resolution. These conditions can arise from patient movement during the scan, technical limitations of the imaging equipment, or the necessity to combine multiple images into a single slice. Despite these inherent challenges, the software consistently delivered accurate segmentations comparable to manual segmentation. This consistency in performance, even with lower quality scans, underscores its potential clinical applicability, where variability in image quality is common but accurate assessments of body composition remain crucial.

*Efficiency of Segmentation*

One of the most significant advantages of using the automated segmentation software is its efficiency. The software completed the segmentation of 423 single-slice L3 MRI images in 126.9 minutes, averaging 18 seconds per DICOM image. In contrast, manual segmentation correction of this dataset on the automatic segmentation took approximately 10-15 minutes per scan after. Manual segmentation from a raw MRI image containing no segmentation would take a significant amount of time, ranging from a few hours per scan. This efficiency gain highlights the potential of automated tools like DAFS Express to streamline workflows and reduce the labor-intensive nature of manual segmentation.

*Clinical Implications and Future Directions*

Our findings support the adoption of automated segmentation tools in clinical and research settings due to their efficiency, reproducibility, and accuracy. By reducing manual labor and variability in segmentations, these tools offer significant advantages in large-scale studies and longitudinal assessments of body composition [4].

Future research directions could explore further enhancements in software algorithms to handle even more diverse imaging challenges and expand applications across different populations and clinical conditions. One potential challenge is the variability in image quality across different scanners and settings. For instance, low-resolution images or images with high levels of noise may affect the accuracy of segmentation algorithms. Another challenge is the presence of imaging artifacts, such as metal implants or motion artifacts, which can distort the images and complicate automated analysis. Furthermore, in oncology patients, tumors and related treatments may alter normal anatomy, thus requiring more sophisticated algorithms to distinguish between normal and pathological tissues.

Moreover, automated segmentation reduces variability between raters, providing more consistent and reproducible results. This consistency is particularly valuable in longitudinal studies, where reliable tracking of changes in body composition over time is crucial. Automated tools ensure that measurements are standardized across different time points and different raters, leading to more reliable data and robust conclusions.

In conclusion, this study contributes valuable insights into the efficacy and reliability of automated segmentation tools for assessing body composition from medical images such as MRI. The combination of robust statistical analyses, visual representations, and assessments under challenging imaging conditions highlights the DAFS Express's potential to advance clinical practice and research in body composition analysis.

**Conclusion**

In this study, DAFS Express, an automated segmentation software, was evaluated for its performance in analyzing 2D MRI images of subcutaneous skeletal muscle (SKM), visceral adipose tissue (VAT), and subcutaneous adipose tissue (SAT). The software demonstrated accuracy comparable to manual methods in delineating tissue areas, showcasing its potential to streamline and enhance image analysis processes in real life settings. Overall, DAFS Express presents a promising tool for advancing automated image analysis, with implications for improving diagnostic accuracy and efficiency in medical imaging research and practice. Continued development and validation will be critical in harnessing its full potential for diverse clinical settings and research applications.

**Supplementary figures**

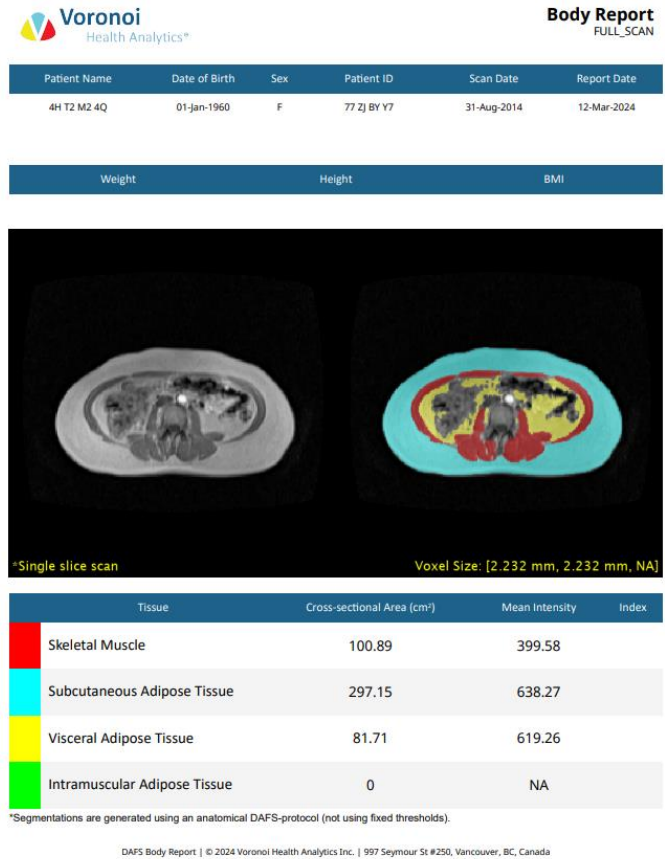

**Supplementary Figure 1** An example PDF report printed by DAFS Express for a single 2D MRI slice at the L3 vertebral level, showcasing the automatic cross-sectional areas (cm$^2$) and mean intensities.